\documentclass[5p, final]{elsarticle}
\usepackage{amsmath}
\usepackage{lscape}
\usepackage{hyperref}
\usepackage{multirow}
\usepackage{booktabs}
\usepackage[noend]{algorithm2e}
\usepackage{enumitem}
\usepackage{url}

\usepackage{subfig}

\usepackage{amssymb}

\usepackage{mathtools}
\newcommand{\defeq}{\vcentcolon=}

\begin{document}
\begin{frontmatter}




\title{Testing the Consistency of Performance Scores Reported for Binary Classification Problems}


\author[1]{Attila Fazekas} 
\ead{attila.fazekas@inf.unideb.hu}

\author[2]{Gy\"orgy Kov\'acs}
\ead{gyuriofkovacs@gmail.com}

\affiliation[1]{organization={Faculty of Informatics, University of Debrecen},
    addressline={Kassai út 26}, 
    city={Debrecen},
    postcode={4028}, 
    country={Hungary}}
\affiliation[2]{organization={Analytical Minds Ltd.},
    addressline={Árpád út 5}, 
    city={Beregsurány},
    postcode={4933}, 
    country={Hungary}}

\begin{abstract}

Binary classification is a fundamental task in machine learning, with applications spanning various scientific domains. Whether scientists are conducting fundamental research or refining practical applications, they typically assess and rank classification techniques based on performance metrics such as \emph{accuracy}, \emph{sensitivity}, and \emph{specificity}. However, reported performance scores may not always serve as a reliable basis for research ranking. This can be attributed to undisclosed or unconventional practices related to cross-validation, typographical errors, and other factors.
In a given experimental setup, with a specific number of positive and negative test items, most performance scores can assume specific, interrelated values. In this paper, we introduce numerical techniques to assess the consistency of reported performance scores and the assumed experimental setup. Importantly, the proposed approach does not rely on statistical inference but uses numerical methods to identify inconsistencies with certainty.
Through three different applications related to medicine, we demonstrate how the proposed techniques can effectively detect inconsistencies, thereby safeguarding the integrity of research fields. To benefit the scientific community, we have made the consistency tests available in an open-source Python package.

\end{abstract}


\end{frontmatter}
        

\thispagestyle{empty}
\section{Introduction}\label{section:Introduction}

Recently, various authors have warned the scientific community about the so-called \emph{reproducibility crisis} in artificial intelligence and machine learning-based research \cite{leakage, reprcrisis, repr0, repr1}. The crisis is characterized by a significant body of research results that prove challenging to replicate through independent experiments or are built upon flawed evaluation methodologies. Among several reasons, notable ones include the absence of shared code \cite{leakage}, improper use of statistics \cite{leakage, staterrors}, 'cosmetic' adjustments of findings \cite{fabrication} and the presence of typographical errors in reported figures.

To address the crisis, various recommendations have been proposed \cite{repr0, repr2} regarding the proper reporting of machine learning research results. Unfortunately, the adoption of these standards has been relatively slow, and they are unable to rectify existing issues in numerous fields.
In practical terms, many domains within machine learning research operate as self-organized, ongoing competitions, with the primary goal of achieving the best outcomes for specific problems and datasets. Within these domains, research quality is often assessed solely based on reported performance scores, despite the proven difficulty of this task even in under controlled circumstances \cite{ranking}. In such circumstances, reliability of reported performance scores becomes crucial. Once published, unrealistically high performance results can inadvertently validate flawed methodologies and may be amplified by publication bias \cite{publicationbias}, ultimately distorting entire research fields.
For example, consider the reports of nearly perfect predictive results for premature delivery in pregnancy based on electrohysterograms (EHG) by multiple authors \cite{ehgreview}. Subsequently, in \cite{ehg}, it was discovered that the overly optimistic results in 11 papers were due to a systematic data leakage in the evaluation methodology.

In general, most meta-analyses \cite{metaresearch} related to reproducibility necessitate manual effort. One approach involves a thorough examination of the relevant papers, seeking deviations from the scientific method and best practices of statistics, as demonstrated in studies such as \cite{psychiatry}, \cite{csecurity}, and \cite{satellite}. Another approach entails attempting to replicate various results by re-implementing the techniques, as exemplified by the authors of the EHG report \cite{ehg}, which demands even more manual labor and becomes impractical at a larger scale.
To facilitate meta-analysis, in this paper, we introduce numerical techniques to test the consistency of reported performance scores with the described experimental setups in binary classification.

Being one of the fundamental tasks in machine learning, binary classification \cite{mlbook} also suffers from the aforementioned problems resulting in the reporting of performance scores that are often incomparable and irreproducible. Whether in fundamental research or practical applications, binary classification performance is typically assessed by making predictions on a test set (sometimes in a cross-validation scheme \cite{cv1}) and constructing the confusion matrix \cite{scores} tabulating the counts of correctly and incorrectly classified positive and negative instances. In practice, confusion matrices are rarely presented directly. Instead, these matrices are usually condensed into multiple numerical metrics (e.g., accuracy and f1-score \cite{scores}), often aggregated across multiple dataset folds or even datasets. The multitude of reported scores quantifies various aspects of performance.

Given the natural constraints imposed by the experimental setup on the confusion matrix (e.g., the sum of all entries must match the cardinality of the test set) and considering that the reported performance scores are interrelated (they cannot take arbitrary values independently), the question arises: \emph{Can the reported performance scores be the result of the experiment?} Mathematically, this question relates to the existence of at least one compatible confusion matrix that aligns with the experimental conditions and yields the reported performance scores. Additional complexity is introduced by rounding and potential aggregations. If the problem proves to be infeasible, any attempts to reproduce the results are destined to fail with certainty. 

Earlier, a technique called \emph{DConfusion} was proposed in \cite{dconfusion} to reconstruct certain characteristics (but not the exact entries) of the confusion matrix from reported scores. \emph{DConfusion} was then effectively applied to various studies involving binary classification \cite{errorsml} to assess the validity of reported scores.
Independently, in a previous paper of ours, we introduced a similar approach to estimate the exact number of pixels used for evaluating retinal vessel segmentation techniques \cite{vesselsegm}. Later on, refining the method, we developed a test that proved sensitive enough to identify a systematic and significant methodological inconsistency in the evaluation of retinal vessel segmentation: an analysis involving more than 100 papers led to the conclusion that the rankings of algorithms in the majority of studies are based on figures that cannot be compared directly \cite{vessel}, despite the authors using the same test set of images for evaluation. Drawing inspiration from the successful application of the concept, this paper generalizes the method further and develops consistency tests for many of the most commonly used performance scores and evaluation schemes in binary classification.
The proposed approach differs from \emph{DConfusion} \cite{dconfusion} in several key ways: (a) while \emph{DConfusion} supports only a limited set of performance scores typically encountered in the field of software fault prediction systems, the proposed technique supports the majority of performance scores used in the literature; (b) \emph{DConfusion} neglects the impact of aggregations, whereas the proposed method addresses the aggregation of scores with mathematical rigor; (c) due to the omission of aggregations and the propagation of rounding errors, \emph{DConfusion} might produce false alarms of inconsistency, whereas the inconsistencies identified by the proposed technique are certain.

We emphasize that the proposed consistency tests are numerical rather than statistical, devoid of any probability of type-I errors (false positives). In other words, inconsistencies are conclusively identified, implying that either the assumed experimental setup or the reported scores are incorrect. Given their robustness and ease of use, we believe that the proposed techniques can serve as effective tools for meta-analysis and contribute to enhancing the reproducibility of machine learning-based science.

The contributions of the paper to the field can be summarized as follows:
\begin{enumerate}
\item For performance scores calculated directly from confusion matrices, we introduce a consistency test that supports 20 of the most commonly used performance scores (without synonyms and complements). We highlight that the test can be easily extended to accommodate further score functions.
\item For experimental setups that involve the averaging of performance scores across dataset folds or multiple datasets, we propose a consistency test that supports four widely used scores: accuracy, sensitivity, specificity, and balanced accuracy.
\item We illustrate the practical application of the proposed techniques in three real-world problems, showcasing its effectiveness in identifying research with flawed evaluation methodologies and unreliable performance scores.
\item To benefit the research community, we have released the proposed tests as part of the open-source Python package \verb|mlscorecheck|. This package is available in the standard PyPI repository or on GitHub at \url{https://github.com/FalseNegativeLab/mlscorecheck}.
\end{enumerate}

The paper is organized as follows: Section \ref{sec:problem} introduces the notation, the concept of binary classification, the performance scores and also outlines the problem of consistency testing. Sections \ref{sec:ind} and \ref{sec:agg} present the proposed techniques for various evaluation schemes and scenarios and Section \ref{sec:applications} demonstrates how the techniques can be applied in practice. Finally, conclusions are drawn in Section \ref{sec:conclusions}.

\section{Problem formulation}
\label{sec:problem}

In this section, we formulate the problem we address and introduce the notations and concepts we will reference throughout the rest of the paper.

In binary classification, typically there is a dataset $\mathcal{D} = \lbrace(\mathbf{x}_1, y_1), \dots, (\mathbf{x}_N, y_N)\rbrace$ consisting of $N$ paired \emph{feature vector}s ($\mathbf{x}_i\in\mathbb{R}^{d}$) and \emph{class label}s ($y_i\in\lbrace 0, 1\rbrace$), the classes labelled as $0$ and $1$ commonly referred as \emph{negative} and \emph{positive} classes, respectively.
The primary objective of binary classification is to use this dataset to infer (train) a function $h$ capable of making predictions about the class label of a previously unseen feature vector $\mathbf{x}\in\mathbb{R}^d$ as $h(\mathbf{x})$. All classification techniques can predict class labels, but many of them (such as decision trees \cite{mlbook}, neural networks \cite{mlbook}, etc.) are designed to approximate the posterior distributions $\mathbb{P}(\mathbf{x}|y=c)$, $c\in\lbrace 0, 1\rbrace$ and derive class labels by maximizing the posterior probability, thereby reducing the probability of misclassification \cite{bayesclassifier}. The choice of which classifier outcome to favor depends on the specific application. For instance, image segmentation \cite{segmentation} requires hard labels indicating whether a pixel belongs to an object, whereas many medical applications prioritize ranking cases by the probability (risk) of a condition \cite{binclasranking}. Consequently, performance measurement varies by the field of application, with two predominant approaches: one quantifies how effectively the posterior probabilities rank items, typically using the AUC score \cite{aucsurvey, auc}; the other assesses the quality of label assignment \cite{scores}. This paper focuses on evaluating the quality of labeling.

To eliminate bias, classifiers must be evaluated using feature vectors and corresponding class labels that were not used for training. In the rest of the paper, we refer them as the \emph{evaluation set}, denoted by $\mathcal{E}$. (We note that in many scenarios the terms \emph{test set} and \emph{validation set} are used synonymously.) For evaluation, the class label $\hat{y} \doteq h(\mathbf{x})$ is predicted for each $(\mathbf{x}, y)\in\mathcal{E}$ and by comparing the corresponding pairs $y$ and $\hat{y}$ the \emph{confusion matrix} (see Table \ref{tptnfpfn}) is constructed, tabulating the integer counts of true positive ($tp$), true negative ($tn$), false positive ($fp$), and false negative ($fn$) predictions. 
One can readily see, that if the number of positive $p$ and negative $n$ instances of the evaluation set $\mathcal{E}$ is known, the binary confusion matrix has two degrees of freedom, since $p = tp + fn$ and $n = tn + fp$. Without loss of generality, we consider $tp$ and $tn$ as the independent components.

To facilitate the comparison of classification approaches, a confusion matrix is often summarized by scalar \emph{performance score}s. (We use the term performance score to prevent confusion with the mathematical notions of \emph{measure} and \emph{metric} which are used synonymously in various sources.) The literature has proposed a multitude of such scores, each emphasizing different aspects of performance, some widely used (such as \emph{accuracy}), while others tailored to specific fields (such as the \emph{diagnostic odds ratio} in medical applications \cite{dor}). See Table \ref{tab:scores} for a summary of the scores covered in this paper. In the rest of the paper, we always assume that there is a set $\mathcal{S}\subseteq\lbrace acc, sens, spec, \dots \rbrace$ (any of the score abbreviations in Table \ref{tab:scores}) of scores reported. For a specific score $s\in\mathcal{S}$, the standardized functional form of the score is denoted by $f_s$, with $f_s(tp, tn, p, n)$ yielding the true, possibly infinite decimal value of the score for a confusion matrix characterized by $tp$, $tn$, $p$ and $n$.

\begin{table}[t!]
\caption{The potential outcomes in the evaluation of binary classification.}
\label{tptnfpfn}
\begin{small}
\begin{center}
\begin{tabular}{c@{\hspace{4pt}}|@{\hspace{4pt}}c@{\hspace{4pt}}c}
& \multicolumn{2}{c}{predicted} \\
& positive ($\hat{y}_i = 1$) & negative ($\hat{y}_i = 0$) \\ 
\hline
positive ($y_i=1)$ & true positive ($tp$) & false negative ($fn$) \\
negative ($y_i=0)$ & false positive ($fp$) & true negative ($tn$) \\
\end{tabular}
\end{center}
\end{small}
\end{table}

In cases where predefined evaluation sets are not available, the common practice is adopting the \emph{hold-out} approach: randomly partitioning the dataset into two disjoint sets (the training set  $\mathcal{T}$, and the evaluation set $\mathcal{E}$).
The random split makes the estimated performance scores uncertain, which can be mitigated by repeating the random partitioning and evaluation multiple times, and aggregating the results. To ensure that all data points are represented equally in the evaluation, usually \emph{k-fold cross-validation} (kFCV) is used \cite{cv1}.
In a kFCV scheme, the dataset $\mathcal{D}$ is randomly divided into $k$ disjoint subsets of equal sizes, referred to as \emph{folds}. The evaluation process iteratively selects one fold as the evaluation set, trains the classifier on the remaining $k-1$ folds, and evaluates it on the selected fold, using all data points for evaluation once. Finally, the results are aggregated over all folds. 
To improve stability further, the kFCV can be repeated multiple times with different random partitionings of $\mathcal{D}$ (known as \emph{repeated kFCV}). Another common enhancement to kFCV is applying \emph{stratification} to ensure that the class distributions in each fold approximate that of the entire dataset. We also note that in many cases, classifiers are evaluated on and the results aggregated over multiple datasets.

Since many performance scores are fractions of integers, it is common practice in scientific writing to round them to a finite number of decimal places for presentation. For instance, introducing the notation $\hat{v}_s$ for the reported figure of the score $s$, $\hat{v}_s = 0.945$ implies that the original value was rounded to $3$ decimal places. Consequently, the true value $v_s^{*} = f_s(tp^{*}, tn^{*}, p, n)$ is expected to fall within the interval $v_s^{*} \in [\hat{v}_s - \epsilon, \hat{v}_s + \epsilon] = [0.9445, 0.9455]$, where $\epsilon = 10^{-3}/2$ represents the \emph{numerical uncertainty}. If flooring and ceiling are also allowed, the numerical uncertainty extends to $\epsilon = 10^{-3}$.

The problem we address in the rest of the paper can be phrased as follows: \emph{Given a set of reported performance scores and the description of the experiment (the datasets involved, the evaluation scheme, the mode of aggregation), could the experiment have yielded the reported scores}?

\begin{table*}
\caption{The table provides a summary of all performance scores discussed in the paper, including their standardized forms that depend on $tp$ and $tn$ only, their original definition, and descriptions that mention common synonyms and complements.}
\label{tab:scores}
\begin{scriptsize}
\begingroup
\renewcommand{\arraystretch}{3.0}
\begin{tabular}{p{0.11474469305794606\textwidth}@{\hspace{2pt}}p{0.045897877223178424\textwidth}@{\hspace{2pt}}p{0.28686173264486514\textwidth}@{\hspace{2pt}}p{0.1950659781985083\textwidth}@{\hspace{2pt}}p{0.3098106712564544\textwidth}}
\toprule
name & abbr. & standardized form & original definition & description \\
\midrule
\parbox{0.11474469305794606\textwidth}{accuracy \cite{scores}} & acc & $\dfrac{tn + tp}{n + p}$ &  & \parbox{0.3098106712564544\textwidth}{The proportion of correctly classified items. Complement: error rate.} \\
\parbox{0.11474469305794606\textwidth}{sensitivity \cite{scores}} & sens & $\dfrac{tp}{p}$ &  & \parbox{0.3098106712564544\textwidth}{The proportion of correctly classified positive items. Also known as: recall, true positive rate. Complement: false negative rate.} \\
\parbox{0.11474469305794606\textwidth}{specificity \cite{scores}} & spec & $\dfrac{tn}{n}$ &  & \parbox{0.3098106712564544\textwidth}{The proportion of correctly classified negative items. Also known as: selectivity, true negative rate. Complement: false positive rate.} \\
\parbox{0.11474469305794606\textwidth}{positive \\ predictive \\ value \cite{scores}} & ppv & $\dfrac{tp}{n - tn + tp}$ &  & \parbox{0.3098106712564544\textwidth}{The proportion of truly positive items among all items classified as positive. Also known as: precision. Complement: false discovery rate.} \\
\parbox{0.11474469305794606\textwidth}{negative \\ predictive \\ value \cite{scores}} & npv & $\dfrac{tn}{p + tn - tp}$ &  & \parbox{0.3098106712564544\textwidth}{The proportion of truly negative items among all items classified as negative. Complement: false omission rate.} \\
\parbox{0.11474469305794606\textwidth}{$f^{\beta}_+$ \cite{scores}} & fbp & $\dfrac{tp \left(\beta_{+}^{2} + 1\right)}{\beta_{+}^{2} p + n - tn + tp}$ &  & \parbox{0.3098106712564544\textwidth}{The harmonic mean of true positive rate and sensitivity, when sensitivity is beta times more important than the true positive rate.} \\
\parbox{0.11474469305794606\textwidth}{$f^{\beta}_-$ \cite{upm}} & fbn & $\dfrac{tn \left(\beta_{-}^{2} + 1\right)}{\beta_{-}^{2} n + p + tn - tp}$ &  & \parbox{0.3098106712564544\textwidth}{The harmonic mean of true negative rate and specificity, when specificity is beta times more important than the true negative rate.} \\
\parbox{0.11474469305794606\textwidth}{unified \\ performance \\ measure \cite{upm}} & upm & $\dfrac{4 tn \cdot tp}{tn \left(n + p - tn + tp\right) + tp \left(n + p + tn - tp\right)}$ & $\dfrac{2 f^{1}_{+} \cdot f^{1}_{-}}{f^{1}_{+} + f^{1}_{-}}$ & \parbox{0.3098106712564544\textwidth}{The harmonic mean of $f^1_-$ and $f^1_+$. Also known as: p4.} \\
\parbox{0.11474469305794606\textwidth}{geometric \\ mean \cite{scores}} & gm & $\sqrt{\dfrac{tn \cdot tp}{np}}$ & $\sqrt{sens \cdot spec}$ & \parbox{0.3098106712564544\textwidth}{The geometric mean of sensitivity and specificity.} \\
\parbox{0.11474469305794606\textwidth}{Fowlkes-Mallows \\ index \cite{fm}} & fm & $\dfrac{tp}{\sqrt{p \left(n - tn + tp\right)}}$ & $\sqrt{ppv \cdot sens}$ & \parbox{0.3098106712564544\textwidth}{The geometric mean of positive predictive value and sensitivity.} \\
\parbox{0.11474469305794606\textwidth}{markedness \cite{mkbm}} & mk & $\dfrac{tn}{p + tn - tp} + \dfrac{tp}{n - tn + tp} - 1$ & $npv + ppv - 1$ & \parbox{0.3098106712564544\textwidth}{The markedness quantifies how marked a condition is for the predictor. Also known as: $\Delta p$.} \\
\parbox{0.11474469305794606\textwidth}{bookmaker \\ informedness \cite{mkbm}} & bm & $-1 + \dfrac{tp}{p} + \dfrac{tn}{n}$ & $sens + spec - 1$ & \parbox{0.3098106712564544\textwidth}{Quantifies how informed the classifier is for the specified condition. Also known as: informedness.} \\
\parbox{0.11474469305794606\textwidth}{Matthews \\ correlation \\ coefficient \cite{scores}} & mcc & $\dfrac{tn \cdot tp - \left(n - tn\right) \left(p - tp\right)}{\sqrt{n p \left(n - tn + tp\right) \left(p + tn - tp\right)}}$ & $\sqrt{bm \cdot mk}$ & \parbox{0.3098106712564544\textwidth}{The Pearson correlation coefficient computed for the observed and predicted labels. Also known as: phi coefficient.} \\
\parbox{0.11474469305794606\textwidth}{positive \\ likelihood \\ ratio \cite{scores}} & lrp & $\dfrac{n \cdot tp}{p \left(n - tn\right)}$ & $\dfrac{sens}{1 - spec}$ & \parbox{0.3098106712564544\textwidth}{Quantifies the probability of correct positive prediction relative to the probability of type I error.} \\
\parbox{0.11474469305794606\textwidth}{negative \\ likelihood \\ ratio \cite{scores}} & lrn & $\dfrac{n \left(p - tp\right)}{p \cdot tn}$ & $\dfrac{1 - sens}{spec}$ & \parbox{0.3098106712564544\textwidth}{Quantifies the probability of type II error relative to the probability of correct negative prediction.} \\
\parbox{0.11474469305794606\textwidth}{prevalence \\ threshold \cite{pt}} & pt & $- \dfrac{p \left(n \sqrt{\dfrac{tp \left(n - tn\right)}{n p}} - n + tn\right)}{- n \cdot tp + p \left(n - tn\right)}$ & $\dfrac{spec + \sqrt{sens \left(1 - spec\right)} - 1}{sens + spec - 1}$ & \parbox{0.3098106712564544\textwidth}{Estimates the threshold on prevalence under which the precision of classification declines rapidly.} \\
\parbox{0.11474469305794606\textwidth}{diagnostic \\ odds \\ ratio \cite{scores}} & dor & $\dfrac{tn \cdot tp}{\left(n - tn\right) \left(p - tp\right)}$ & $\dfrac{lr_{+}}{lr_{-}}$ & \parbox{0.3098106712564544\textwidth}{The ratio of the odds that tha classifier correctly predicts a positive label to the odds of incorrectly predicting the positive label.} \\
\parbox{0.11474469305794606\textwidth}{Jaccard \\ index \cite{scores}} & ji & $\dfrac{tp}{n + p - tn}$ &  & \parbox{0.3098106712564544\textwidth}{The intersection over the union respecting items predicted positive and observed positive. Also known as: threat score, ratio of verification, critical success index, Tanimoto coefficient.} \\
\parbox{0.11474469305794606\textwidth}{balanced \\ accuracy \cite{scores}} & bacc & $\dfrac{tp}{2 p} + \dfrac{tn}{2 n}$ & $\dfrac{sens}{2} + \dfrac{spec}{2}$ & \parbox{0.3098106712564544\textwidth}{The mean of sensitivity and specificity.} \\
\parbox{0.11474469305794606\textwidth}{Cohen's \\ kappa \cite{kappa}} & kappa & $\dfrac{- 2 n p + 2 n \cdot tp + 2 p \cdot tn}{n^{2} - n \cdot tn + n \cdot tp + p^{2} + p \cdot tn - p \cdot tp}$ &  & \parbox{0.3098106712564544\textwidth}{Quantifies the agreement between the observed labeling and the labeling by the classifier, taking into account the probability of agreement by chance.} \\
\bottomrule
\end{tabular}
\endgroup
\end{scriptsize}
\end{table*}

\section{Testing scores derived from one confusion matrix}
\label{sec:ind}

In this section, we assume that there is a well-specified evaluation set available, with known numbers of positive ($p$) and negative ($n$) instances, and a set of scores $\mathcal{S}\subseteq\lbrace acc, sens, \dots\rbrace$ reported up to $\epsilon$ numerical uncertainty. 
It is worth noting that this experimental setup is common in various fields such as computer vision (when results for publicly available test images are shared \cite{vessel, isic2016}); big data (where the hold-out strategy is used to reduce computational demand); and machine learning competitions, including those on platforms like Kaggle (\url{www.kaggle.com}), where a closed test set is withheld.

The section is organized as follows: in Section \ref{sec:rommat} we introduce the consistency test in terms of exhaustive search, in Section \ref{sec:improved} we propose a more efficient implementation using interval computing and finally, the method is illustrated through an example in Section \ref{sec:index}.

\subsection{Testing by exhaustive search}
\label{sec:rommat}

The experimental setup (described by $p$ and $n$) and the reported scores are consistent if there exist some $tp^* \in \mathcal{P} \doteq \lbrace 0, \dots, p\rbrace$ and $tn^*\in\mathcal{N}\doteq \lbrace 0, \dots, n\rbrace$ integers, such that
\begin{equation}
\label{eqtest0}
f_s(tp^*, tn^*, p, n) \in [\hat{v}_s - \epsilon, \hat{v}_s + \epsilon],
\end{equation}
holds for each score $s\in\mathcal{S}$ simultaneously. This simple condition readily suggests an $O(p\cdot n\vert S\vert)$ time complexity algorithm based on exhaustive search: 
one can test each pair of $(tp, tn)\in \mathcal{P}\times\mathcal{N}$ if they satisfy the conditions. If there are no feasible pairs, the experimental setup and the reported scores are inconsistent. Although this brute force algorithm is functional and can be applied to datasets of even medium sizes ($p + n \lessapprox 10K$), it is possible to reduce its time complexity making it efficient for much larger datasets.

\subsection{The improved time complexity test}
\label{sec:improved}

The idea behind the improvement is that the iteration through the potential values of both $tp$ and $tn$ could be eliminated if we could determine for a given value of $tp$ the values of $tn$ leading to a desired performance score $v_s^{*}$. To do so, the analytical forms of the score functions need to be solved for the particular variables. Namely, from $v_s = f_s(tp, tn, p, n)$ solving
\begin{equation}
v_s - f_s(tp, tn, p, n) = 0
\end{equation}
for $tn$ leads to the solution
\begin{align}
\label{eq:solution}
tn &= f_{s, tn}^{-1}(v_s, tp, p, n).
\end{align}
The solutions for all scores are provided in tables \ref{tab2} and \ref{tab3}.

If we knew the exact values of the scores $v_s^*$, one would need to check if there exist any $tp\in\mathcal{P}$ such that \break $f_{s, tn}^{-1}(v_s^*, tp, p, n)$ gives the same integer result for each $s\in\mathcal{S}$.

In practice, we have only interval estimations for $v_s^* \in [\hat{v}_s - \epsilon, \hat{v}_s + \epsilon]$. Therefore, to evaluate (\ref{eq:solution}) one needs to exploit interval arithmetics \cite{interval}. For example, given $p=40$, $n=70$ and $\hat{v}_{acc} = 0.927$, one wants to determine which $tn$ values could lead to the reported score if $tp=30$ (an arbitrary choice from $\mathcal{P}$). Selecting the solution $f^{-1}_{acc, tn}$ from Table \ref{tab2} and evaluating it with interval arithmetics yields
\begin{equation}
f^{-1}_{acc, tn}([0.926, 0.928], 30, 40, 70) = [71.86, 72.08],
\end{equation}
that is, the only integer $tn$ can take to result the reported accuracy score is $tn=72$. We note that depending on the numerical uncertainty, the resulting intervals might contain multiple integers and some scores have multiple solutions leading the the union of intervals (Table \ref{tab3}).

Ultimately, the consistency test can be carried out by iteratively testing if there exist $tp\in\mathcal{P}$  such that the intersection $f_{s, tn}^{-1}([\hat{v}_s - \epsilon, \hat{v}_s + \epsilon], tp, p, n)$ for all $s\in\mathcal{S}$ contains at least one integer in the feasible set $\mathcal{N}$. If no such $tp\in\mathcal{P}$ exists, the experimental setup and the reported scores are inconsistent with each other. 
One can readily see, that the choice of $tp$ to iterate by is arbitrary, the consistency test could be implemented by iterating by $tn\in\mathcal{N}$ and using solutions for $tp$ (also given in table \ref{tab2} and \ref{tab3}). Consequently, the time complexity can be reduced to $O(\min(p, n)\cdot \vert S\vert)$ if the figure with the smaller domain is chosen for iteration, leading to an efficient, linear time algorithm which is tractable even when the evaluation set contains millions of records. The pseudo-code of the test is listed in Algorithm \ref{alg1}.

We note that the sensitivity of the test (ability to recognize inconsistencies) is dependent on the specifics of the experiment, and the number and precision of reported scores, however as we demonstrate in Section \ref{sec:applications}, it can be applied successfully in many typical scenarios.

It is worth noting that the time complexity could be further reduced by solving pairs of performance scores as a system for $tp$ and $tn$, eliminating the need for iteration by $tp$ or $tn$. However, we found that solving pairs of scores with higher order terms would require the involvement of advanced algebraic techniques, which falls beyond the scope of this paper.

\SetKwComment{Comment}{/* }{ */}

\begin{algorithm}[t]
\caption{Consistency testing for scores computed directly from the confusion matrix}\label{alg1}
\begin{small}
\KwData{$p$, $n$, $\epsilon$, the set of scores reported $\mathcal{S}$, the reported values $\hat{v}_s$, $s\in\mathcal{S}$}
\KwResult{$True$ if inconsistency was found, $False$ otherwise.}
\Comment{Selecting the figure to solve for ($\beta$), the feasible set ($\mathcal{B}$), and the domain to iterate on ($\mathcal{A})$}
\eIf{$p < $n}{
$\mathcal{A} \gets \lbrace 0, \dots, p\rbrace$\; 
$\mathcal{B} \gets \lbrace 0, \dots, n\rbrace$\; 
$\beta \gets \text{'tn'}$\;
}{
$\mathcal{A} \gets \lbrace 0, \dots, n\rbrace$\;
$\mathcal{B} \gets \lbrace 0, \dots, p\rbrace$\;
$\beta \gets \text{'tp'}$\;
}
\Comment{Iterate through the possible values}
\For{$\alpha \in \mathcal{A}$}{
    $I \gets \bigcap\limits_{s\in\mathcal{S}} f^{-1}_{s, \beta}([\hat{v}_s-\epsilon, \hat{v}_s+\epsilon], \alpha, p, n)$\;
    \If{$I \bigcap \mathcal{B}$ contains an integer}{
     \Comment{Evidence found for feasibility}
      \Return False\;
    }
}
\Comment{Inconsistency identified}
\Return True\;
\end{small}
\end{algorithm}

\begin{table*}[t!]
\caption{Scores with single solutions.}
\label{tab2}
\begin{scriptsize}
\begin{center}
\begin{tabular}{lll}
\toprule
score ($s$) & tp ($f^{-1}_{s, tp}(s, tn, p, n)$) & tn ($f^{-1}_{s, tn}(s, tp, p, n)$)\\
\midrule
acc & $acc \cdot n + acc \cdot p - tn$ & $acc \cdot n + acc \cdot p - tp$ \\
sens & $p \cdot sens$ &  \\
spec &  & $n \cdot spec$ \\
ppv & $\dfrac{ppv \left(- n + tn\right)}{ppv - 1}$ & $n + tp - \dfrac{tp}{ppv}$ \\
npv & $p + tn - \dfrac{tn}{npv}$ & $\dfrac{npv \left(- p + tp\right)}{npv - 1}$ \\
fbp & $\dfrac{f^\beta_{+} \left(\beta_{+}^{2} p + n - tn\right)}{\beta_{+}^{2} - f^\beta_{+} + 1}$ & $\dfrac{- \beta_{+}^{2} tp + f^\beta_{+} \left(\beta_{+}^{2} p + n + tp\right) - tp}{f^\beta_{+}}$ \\
fbn & $\dfrac{- \beta_{-}^{2} tn + f^\beta_{-} \left(\beta_{-}^{2} n + p + tn\right) - tn}{f^\beta_{-}}$ & $\dfrac{f^\beta_{-} \left(\beta_{-}^{2} n + p - tp\right)}{\beta_{-}^{2} - f^\beta_{-} + 1}$ \\
gm & $\dfrac{gm^{2} n p}{tn}$ & $\dfrac{gm^{2} n p}{tp}$ \\
lrp & $\dfrac{lr_{+} p \left(n - tn\right)}{n}$ & $n - \dfrac{n \cdot tp}{lr_{+} p}$ \\
lrn & $\dfrac{p \left(- lr_{-} tn + n\right)}{n}$ & $\dfrac{n \left(p - tp\right)}{lr_{-} p}$ \\
bm & $\dfrac{p \left(n \left(bm + 1\right) - tn\right)}{n}$ & $\dfrac{n \left(p \left(bm + 1\right) - tp\right)}{p}$ \\
pt & $\dfrac{p \left(n - tn\right)}{n}$ & $\dfrac{n \left(p - tp\right)}{p}$ \\
dor & $\dfrac{dor \cdot p \left(n - tn\right)}{dor \cdot n - dor \cdot tn + tn}$ & $\dfrac{dor \cdot n \left(p - tp\right)}{dor \cdot p - dor \cdot tp + tp}$ \\
ji & $ji \left(n + p - tn\right)$ & $n + p - \dfrac{tp}{ji}$ \\
bacc & $\dfrac{p \left(2 bacc \cdot n - tn\right)}{n}$ & $\dfrac{n \left(2 bacc \cdot p - tp\right)}{p}$ \\
kappa & $\dfrac{\kappa \left(n^{2} - n \cdot tn + p^{2} + p \cdot tn\right) + 2 n p - 2 p \cdot tn}{\kappa \left(- n + p\right) + 2 n}$ & $\dfrac{\kappa \left(n^{2} + n \cdot tp + p^{2} - p \cdot tp\right) + 2 n p - 2 n \cdot tp}{\kappa \left(n - p\right) + 2 p}$ \\
\bottomrule
\end{tabular}
\end{center}
\end{scriptsize}
\end{table*}

\begin{table*}[t!]
\caption{Scores with multiple solutions.}
\label{tab3}
\begin{scriptsize}
\begin{center}
\begin{tabular}{lll}
\toprule
score & variable & value \\
\midrule
\multirow{2}{*}{fm} & tp$_{1,2}$ & $\dfrac{fm \left(fm \cdot p \pm \sqrt{p} \sqrt{fm^{2} p + 4 n - 4 tn}\right)}{2}$ \\
 & tn & $n + tp - \dfrac{tp^{2}}{fm^{2} p}$ \\
\multirow{2}{*}{mcc} & tp$_{1,2}$ & $\dfrac{\pm mcc \sqrt{p} \left(n + p\right) \sqrt{mcc^{2} n p + 4 n \cdot tn - 4 tn^{2}} + \sqrt{n} p \left(mcc^{2} \left(- n + p + 2 tn\right) + 2 n - 2 tn\right)}{2 \sqrt{n} \left(mcc^{2} p + n\right)}$ \\
 & tn$_{1,2}$ & $\dfrac{\pm mcc \sqrt{n} \left(n + p\right) \sqrt{mcc^{2} n p + 4 p \cdot tp - 4 tp^{2}} + n \sqrt{p} \left(mcc^{2} \left(n - p + 2 tp\right) + 2 p - 2 tp\right)}{2 \sqrt{p} \left(mcc^{2} n + p\right)}$ \\
\multirow{2}{*}{mk} & tp$_{1,2}$ & $\dfrac{mk \left(- n + p + 2 tn\right) - n \pm \sqrt{mk^{2} \left(n^{2} + 2 n p + p^{2}\right) + mk \left(2 n^{2} + 2 n p - 4 n \cdot tn - 4 p \cdot tn\right) + n^{2}}}{2 mk}$ \\
 & tn$_{1,2}$ & $\dfrac{mk \left(n - p + 2 tp\right) \pm p - \sqrt{mk^{2} \left(n^{2} + 2 n p + p^{2}\right) + mk \left(2 n p - 4 n \cdot tp + 2 p^{2} - 4 p \cdot tp\right) + p^{2}}}{2 mk}$ \\
\multirow{2}{*}{upm} & tp$_{1,2}$ & $\dfrac{n}{2} + \dfrac{p}{2} + tn + \dfrac{- 2 tn \pm \dfrac{\sqrt{16 tn^{2} + upm^{2} \left(n^{2} + 2 n p + 8 n \cdot tn + p^{2} + 8 p \cdot tn\right) + upm \left(- 8 n \cdot tn - 8 p \cdot tn - 16 tn^{2}\right)}}{2}}{upm}$ \\
 & tn$_{1,2}$ & $\dfrac{n}{2} + \dfrac{p}{2} + tp + \dfrac{- 2 tp \pm \dfrac{\sqrt{16 tp^{2} + upm^{2} \left(n^{2} + 2 n p + 8 n \cdot tp + p^{2} + 8 p \cdot tp\right) + upm \left(- 8 n \cdot tp - 8 p \cdot tp - 16 tp^{2}\right)}}{2}}{upm}$ \\
\bottomrule
\end{tabular}
\end{center}
\end{scriptsize}
\end{table*}

\subsection{Example}
\label{sec:index}

Suppose, there is an evaluation set of $p=1000$ positive and $n=6000$ negative samples, and the reported score $\hat{v}_{acc} = 0.6801$, $\hat{v}_{npv} = 0.9401$ and $\hat{v}_{f_1} = 0.4004$. Being conservative and assuming the scores are floored or ceiled, the numerical uncertainty is $\epsilon = 0.0001$. Applying algorithm \ref{alg1}, one finds that there are two pairs of ($tp$, $tn$) values, compatible with the setup: (743, 4031); (743, 4032). If the scores were adjusted a little, for example, accuracy changed to $0.6811$, there are no ($tp$, $tn$) pairs fulfilling all conditions. Similarly, if the assumption on $p$ was different, for example, $p=1100$, the scores turn to be inconsistent.

\section{Testing scores derived by aggregations}
\label{sec:agg}

In this section, we develop tests for those scenarios when the scores are aggregated over multiple evaluation sets (folds and/or datasets). 
The mode of aggregation (discussed in Section \ref{sec:rommor}) leads to different tests that we cover in Sections \ref{sec:rom} and \ref{sec:mor}. 
Finally, the mapping of some common kFCV schemes to the representation used by the tests is discussed in Section \ref{sec:mapping}. 

\subsection{Mean of Scores and Score of Means aggregations}
\label{sec:rommor}

We assume an experiment involving $N_e$ evaluation sets with $p_i$ and $n_i$, $i=1,\dots,N_e$ positive and negative samples, respectively, each leading to a separate confusion matrix with entries $tp_i \in\lbrace 0, \dots, p_i\rbrace$ and $tn_i\in\lbrace 0, \dots, n_i\rbrace$. We are concerned about how these figures are summarized by scalar scores describing the entire experiment. 

A natural way of aggregation is to calculate the scores for each evaluation set, and take the averages of the scores. Formally, for a particular score $s$, the overall score is calculated as
\begin{equation}
\label{estmor}
v_s^{MoS} = \dfrac{1}{N_e}\sum\limits_{i=1}^{N_e} f_s(tp_{i}, tn_{i}, p_{i}, n_{i}),
\end{equation}
where we introduced the notion of \emph{Mean of Scores} (MoS) to indicate the way of aggregation. We note that the MoS mode of aggregation is extremely common in the evaluation of binary classifiers in cross-validation scenarios, with the benefit that the $N_e$-sized sample of scores enables the estimation of confidence intervals \cite{morex2} and the use of  hypothesis testing for the comparison of classification techniques \cite{morex0}.

Alternatively, one can calculate the averages of the $tp$, $tn$, $fp$ and $fn$ figures first, for example, $\overline{tp} = \dfrac{1}{N_e}\sum\limits_{i=1}^{N_e} tp_i$, and compute the scores from the mean figures as
\begin{equation}
\label{estrom}
v_s^{SoM} = f_s\left(\overline{tp}, \overline{tn}, \overline{p}, \overline{n}\right),
\end{equation}
where we introduced the notion of \emph{Score of Means} (SoM) to reflect the way of aggregation. One can readily see, that the SoM aggregation is equivalent to a weighted MoS aggregation, when the weights are defined as the denominators of the scores. SoM aggregation is beneficial when the scores for some individual evaluation sets might become undefined, typically with small and imbalanced data \cite{romex0} (for example, if a fold has only a handful of positive samples, $tp=0$ and $fp=0$ leads to an undefined positive predictive value, which is a less likely scenario for $\overline{tp}$ and $\overline{fp}$).

The terms used for the aggregations are inspired by the analogous concepts of \emph{Ratio of Means} (RoM) and \emph{Mean of Ratios} (MoR) estimations for ratio statistics \cite{rommor, rommor2}, but generalized to accommodate the non-linearities in the numerators and denominators of some scores (like Matthews correlation coefficient).

From the theoretical point of view, the goal of using multiple evaluation sets and aggregating the results is to get a more reliable estimation of performance for the population of problems represented by the evaluation sets. (We note that this concept leads to difficulties when multiple datasets are involved, as the population of classification problems represented by some datasets is usually not well-defined \cite{nfl}.) Nevertheless, estimation theory \cite{estimationtheory} can be expected to provide a guideline on which aggregation is more reasonable. Interestingly, already for the simplest scores (with linear terms in the numerator and denominator) it turns out that both aggregation schemes (\ref{estmor}) and (\ref{estrom}) are biased estimators of the population level statistics \cite{rommor2}. Moreover, even in the same experiment, different scores can lead to different interpretations regarding their meaning. For example, in kFCV, if the total number of samples ($p + n$) is divisible by the number of folds, the fold-level accuracy scores have the same constant denominator, the MoS and SoM aggregations become the same, and the aggregated accuracy becomes an unbiased estimator of the population level proportion of correctly classified items. In the same scenario, sensitivity has randomness in the denominator since the various folds can have varying number of positive samples, consequently, one can argue that weighting by the number of positive samples in a fold (using SoM) is a meaningful way to reduce noise. Finally, positive predictive value has correlated randomness in its numerator and denominator (through the presence of $tp$) leading to both the SoM and MoS aggregations becoming biased estimators \cite{rommor2}. Consequently, there is no consensus on the superiority of either mode of aggregation.

In practice, when the data distribution across evaluation sets is fairly uniform, the scores calculated by both aggregations are nearly identical (see Table \ref{mossomex}). Therefore, authors often do not explicitly describe the method of aggregation, as it is not expected to significantly alter the qualitative outcome of the research. A particular choice could be motivated by multiple factors, for example: the best practices of a field; the available implementation; the need to estimate the uncertainty of the scores; small and imbalanced data, etc. Even in the same domain, with the same data, one can find examples of both aggregations \cite{vessel}, therefore, unless explicitly phrased, one cannot assume any aggregation as default.

Since we develop sharp tests to check the consistency of reported scores, even minor differences must be handled with mathematical rigour. Therefore, in Sections \ref{sec:rom} and \ref{sec:mor} we develop consistency tests for the two types of aggregations separately.

\begin{table*}
\caption{Comparison of the MoS and SoM evaluations on sample data in a k-fold cross-validation scenario with $k=5$: the folds and a sample evaluation \ref{tab4a}; and the scores \ref{tab4b}. According to the expectations, the more non-linearities are present in a score, the more the two aggregations deviate, but the most commonly used scores (accuracy, sensitivity, specificity, f$^1$) are very close to each other.}
\label{mossomex}
\begin{center}
\begin{footnotesize}
\subfloat[(a)][The folds.\label{tab4a}]{
\begin{tabular}[t]{r@{\hspace{5pt}}r@{\hspace{5pt}}r@{\hspace{5pt}}r@{\hspace{5pt}}r}
\toprule
fold ($i$) & $p_i$ & $n_i$ & $tp_i$ & $tn_i$ \\
\midrule
0 & 100 & 201 & 78 & 189 \\
1 & 100 & 200 & 65 & 191 \\
2 & 100 & 200 & 81 & 160 \\
3 & 101 & 200 & 75 & 164 \\
4 & 101 & 200 & 72 & 171 \\
\bottomrule
\end{tabular}
}
\subfloat[(b)][The scores calculated by the two aggregation techniques.\label{tab4b}]{
\begin{tabular}{l@{\hspace{5pt}}r@{\hspace{5pt}}r@{\hspace{5pt}}|l@{\hspace{5pt}}r@{\hspace{5pt}}r@{\hspace{5pt}}|l@{\hspace{5pt}}r@{\hspace{5pt}}r@{\hspace{5pt}}|l@{\hspace{5pt}}r@{\hspace{5pt}}r}
\toprule
score & MoS & SoM & score & MoS & SoM & score & MoS & SoM & score & MoS & SoM \\
\midrule
acc & 0.8290 & 0.8290 & $f^1_+$ & 0.7443 & 0.7427 & lrn & 0.2975 & 0.2985 & ppv & 0.7606 & 0.7465 \\
bacc & 0.8066 & 0.8066 & fm & 0.7471 & 0.7428 & lrp & 8.1202 & 5.8713 & pt & 0.2795 & 0.2921 \\
bm & 0.6131 & 0.6132 & gm & 0.8021 & 0.8038 & mcc & 0.6215 & 0.6147 & sens & 0.7391 & 0.7390 \\
dor & 28.0174 & 19.6671 & ji & 0.5945 & 0.5908 & mk & 0.6312 & 0.6163 & spec & 0.8741 & 0.8741 \\
$f^1_-$ & 0.8709 & 0.8719 & kappa & 0.6165 & 0.6147 & npv & 0.8706 & 0.8698 & upm & 0.8025 & 0.8022 \\
\bottomrule
\end{tabular}
}
\end{footnotesize}
\end{center}
\end{table*}

\subsection{Testing scores aggregated by the \emph{Score of Means} approach}
\label{sec:rom}

While we introduced the term \emph{Score of Means} to reflect the analogy with the concept of \emph{Ratio of Means} in statistics, taking the mean of the figures $tp$, $tn$, $p$, and $n$ (equation \ref{estrom}) is unnecessary. It can be readily seen that all scores covered in the paper (Table \ref{tab:scores}) are invariant to scaling. In other words, for any score $s$, the equation $f_s(tp, tn, p, n) = f_s(\alpha\cdot tp, \alpha\cdot tn, \alpha\cdot p, \alpha\cdot n)$ holds for $\alpha \in\mathbb{R}^{+}$, and consequently,
\begin{equation}
f_s(\overline{tp}, \overline{tn}, \overline{p}, \overline{n}) = f_s\left(\sum\limits_{i=1}^{N_e} tp_i, \sum\limits_{i=1}^{N_e} tn_i, \sum\limits_{i=1}^{N_e} p_i, \sum\limits_{i=1}^{N_e} n_i\right).
\end{equation}
Therefore, any score calculated using the SoM approach can be treated as if it was calculated from the confusion matrix of a problem with $p'=\sum\limits_{i=1}^{N_e} p_i$ positive and $n'=\sum\limits_{i=1}^{N_e} n_i$ negative samples. Consequently, when scores aggregated in the SoM manner are reported, the consistency tests developed in Section \ref{sec:ind} are applicable using the total number of positives $p'$ and negatives $n'$.

\subsection{Testing scores aggregated by the \emph{Mean of Scores} approach}
\label{sec:mor}

In this section, we develop consistency tests for the MoS aggregation. First, we formulate the problem mathematically in Section \ref{sec:mosmath}, then propose a tractable algorithm based on linear integer programming in Section \ref{sec:moslinprog}, finally, the use of the technique is illustrated in Section \ref{sec:mosex}.

\subsubsection{Mathematical formulation}
\label{sec:mosmath}
According to the definition of MoS aggregation (equation (\ref{estmor})), we are concerned with the simultaneous feasibility of the inequalities:
\begin{equation}
\label{nlopt}
\hat{v}_s^{MoS} - \epsilon \leq \dfrac{1}{N_e}\sum\limits_{i=1}^{N_e} f_s(tp_i, tn_i, p_i, n_i) \leq \hat{v}_s^{MoS} + \epsilon, \; \text{for } s \in \mathcal{S},
\end{equation}
where $tp_i \in \lbrace 0, \dots, p_i\rbrace$ and $tn_i \in \lbrace 0, \dots, n_i\rbrace$ for $i = 1, \dots, N_e$. Due to the non-linearities and the presence of raw figures $tp_i$ and $tn_i$ in both the numerators and denominators of most scores, the averaging cannot be simplified, resulting in a total of $2N_e$ degrees of freedom in the general case.

Similarly to the approach introduced in Section \ref{sec:rommat}, one could enumerate all possible combinations of the $0 \leq tp_i \leq p_i$ and $0 \leq tn_i \leq n_i$, $i=1, \dots, N_e$ figures and check if any of them leads to the reported scores $\hat{v}_s^{MoS}$, $s\in\mathcal{S}$ within the numerical uncertainty $\epsilon$. However, the time complexity $O\left(\prod_{i=1}^{N_e}p_in_i\right)$ of this brute force approach renders it intractable in practice, even in the simplest cases: a 5-fold evaluation with $p_i\sim 10$ positive and $n_i\sim 10$ negative records in each fold leads to approximately $10^{10}$ different combinations of the free parameters. 

\subsubsection{Feasibility by linear integer programming}
\label{sec:moslinprog}

The condition set (\ref{nlopt}) can be interpreted as the definition of the feasibility region of a non-linear integer programming problem (with any dummy objective function). In general, non-linear integer programming is NP-complete \cite{ip}, with no efficient algorithms for exact solutions and approximations are not suitable for sharp consistency tests requiring exact decisions regarding feasibility. 

On the other hand, for that subset of scores which leads to linear conditions, linear integer programming can be exploited, which is solvable by numerous techniques \cite{ip}. Consequently, the proposed consistency test for MoS aggregations supports only those scores which are linear functions of $tp$ and $tn$, namely, \emph{accuracy}, \emph{sensitivity}, \emph{specificity} and \emph{balanced accuracy}. Although this test is limited to these four scores only, we note that these scores are among the most commonly reported ones. 
Expanding (\ref{nlopt}) for the linear scores leads to the condition set
\begin{align}
\label{aggtest}
\hat{v}_{acc}^{MoS} - \epsilon & \leq \dfrac{1}{N_e} \sum\limits_{i=1}^{N_e} \dfrac{tp_i + tn_i}{p_i + n_i} \leq \hat{v}_{acc}^{MoS} + \epsilon, \nonumber \\
\hat{v}_{sens}^{MoS} - \epsilon & \leq \dfrac{1}{N_e} \sum\limits_{i=1}^{N_e} \dfrac{tp_i}{p_i} \leq \hat{v}_{sens}^{MoS} + \epsilon, \nonumber \\
\hat{v}_{spec}^{MoS} - \epsilon & \leq \dfrac{1}{N_e} \sum\limits_{i=1}^{N_e} \dfrac{tn_i}{n_i} \leq \hat{v}_{spec}^{MoS} + \epsilon, \nonumber \\
\hat{v}_{bacc}^{MoS} - \epsilon & \leq \dfrac{1}{N_e} \sum\limits_{i=1}^{N_e} \dfrac{tp_i}{2p_i} + \dfrac{tn_i}{2n_i} \leq \hat{v}_{bacc}^{MoS} + \epsilon, \nonumber \\
tp_i &\in \lbrace 0, \dots, p_i\rbrace, \quad tn_i \in \lbrace 0, \dots, n_i\rbrace,
\end{align}
which is the most general set of conditions that is compatible with linear integer programming. The consistency test operates by specifying the conditions (\ref{aggtest}) for the available scores $\mathcal{S} \bigcap \lbrace acc, sens, spec, bacc\rbrace$, and using any linear integer programming solver to check the feasibility of the condition set. If the conditions are feasible, there is no inconsistency between the scores and the experimental setup; if the feasibility region is empty, the reported scores and the assumptions on the experimental setup are inconsistent.

The sensitivity of the test (the ability to recognize inconsistencies) highly depends on the structure of the evaluation sets and the numerical uncertainty of the reported scores. However, as we illustrate in Section \ref{sec:applications}, it is able recognize inconsistencies in real scenarios. We also mention that there is one more piece of information that is sometimes reported and can strenghten the test: the minimum and maximum scores across folds. As noted in Section \ref{sec:rommor}, one benefit of using MoS aggregation is that one gets a distribution of scores, and sometimes authors report the minimum and maximum values achieved across the folds. Adding these constraints shrinks the feasibility region and improves the sensitivity of the test. For example, if minimum ($\hat{v}_{min(acc)}^{MoS}$)  and maximum ($\hat{v}_{max(acc)}^{MoS}$) scores are reported for accuracy, additional $N_e$ pieces of constraints can be added to the linear programming problem:

\begin{align}
\hat{v}_{min(acc)}^{MoS} - \epsilon \leq \dfrac{tp_i + tn_i}{p_i + n_i} \leq \hat{v}_{max(acc)}^{MoS} + \epsilon, i = 1, \dots, N_e
\end{align}

We note that under special circumstances (stratified kFCV, both $p$ and $n$ divisible by the number of folds), for some subsets of the scores possibly fractional or convex programming with certain relaxation techniques could be exploited \cite{nonlinear}. However, the exploration of these special cases is beyond the scope of the paper.

\subsubsection{Example}
\label{sec:mosex}

The usage of the test is illustrated through the sample problem shared in Table \ref{mossomex}. Suppose the scores
\begin{equation}
\hat{v}_{acc}^{MoS} = 0.8290, \quad
\hat{v}_{sens}^{MoS} = 0.7391, \quad
\hat{v}_{spec}^{MoS} = 0.8741
\end{equation}
are reported. With a conservative choice of numerical uncertainty being $\epsilon=0.0001$ (allowing ceiling or flooring), substituting the scores and the specifications of the folds from Table \ref{tab4a} into (\ref{aggtest}), and checking the feasibility by a linear integer programming solver (we used the Python package \verb|pulp| \cite{pulp}), the solver returns that the problem is feasible, indicating that the scores could be the outcome of the experiment. However, if accuracy is incorrectly reported as $\hat{v}_{acc}^{MoS} = 0.8280$, the solver returns that the configuration is infeasible, indicating that the scores are incompatible with the assumptions on the experiment.

\subsection{Application of the tests in various experimental setups}
\label{sec:mapping}

In the preceding sections, we introduced consistency tests for two modes of aggregations in terms of $N_e$ evaluation sets. In this section, we discuss how various kFCV experiments can be assessed using these tests. In a kFCV experiment with \emph{k} folds on a dataset comprising \emph{p} positive and \emph{n} negative entries, we define \emph{fold configuration} as the distribution of positives and negatives across the \emph{k} folds, denoted as $(p_i, n_i)$, $i=1, \dots, k$. We assume $p_i + n_i$ is either $\lfloor (p + n) / k\rfloor$ or $\lfloor (p + n) / k\rfloor + 1$ and there are at least two folds containing at least one negative and two folds containing at least one positive sample -- this is a necessary requirement to ensure all training sets in the folding process contain samples of both classes and at least the accuracy scores is computable on each fold.
We treat fold configurations as multisets, a certain pair of positive and negative counts can appear multiple times, but the order of the pairs is irrelevant as the lead to the same linear programming problem regardless of order.

\subsubsection{Testing the SoM scores of kFCV experiments}
\label{sec:somkfcv}

In the case of SoM aggregation, as discussed in Section \ref{sec:rom}, only the overall counts of positive and negative samples are needed for the testing. Therefore, in an ordinary kFCV experiment (evaluating each entry of a dataset once), the parameters of the dataset need to be used. In a repeated kFCV scenario, involving $N_d$ datasets with $N_r$ repetitions, $p'=N_r\cdot \sum\limits_{i=1}^{N_d} p_i$ and $n'=N_r\cdot \sum\limits_{i=1}^{N_d} n_i$ need to be used, the fold configurations are irrelevant.

\subsubsection{Testing the MoS scores of kFCV experiments}
\label{sec:moskfcv}

In the case of MoS aggregations, the fold configuration needs to be known to formulate the linear programming problem (\ref{aggtest}). Although there are some data providers that also supply foldings of the data (for example, the KEEL data repository \cite{keel}), in a general kFCV scenario, the fold configuration is unknown. There are however special cases, when the fold configuration can be inferred. If kFCV is carried out in a stratified manner, the stratification technique can be used to infer the configuration of folds. For example, the stratification implemented in the \verb|sklearn| \cite{sklearn} Python package ensures that folds differ at most in 1 sample regarding the overall count of items, as well as the counts of positives and negatives. This requirement leads to a unique configuration that can be inferred from $p$, $n$ and $k$ which is shared in Table \ref{tab:strfolds}. In the case of repeated kFCV experiments or the use of multiple datasets, the union of all fold configurations need to be used.

\begin{table*}
    \caption{Fold configurations by the stratified kFCV implemented in sklearn with $p_{mod} = p \mod k$, $p_{div} = \lfloor p/k\rfloor$, $n_{mod} = n \mod k$, $n_{div} = \lfloor n/k\rfloor$. The 'count of folds' column indicates how many times folds with $p_i$ and $n_i$ counts appear in the configuration, when $p_{mod} + n_{mod} > k$ (a) and $p_{mod} + n_{mod} \leq k$ (b).}
    \label{tab:strfolds}
\begin{center}
\begin{small}
 \begin{tabular}{cc}   
    \subfloat[][If $p_{mod} + n_{mod} > k$]{
    \begin{tabular}{ccc}
    \toprule
    count of folds & positives ($p_i$) & negatives ($n_i$) \\
    \midrule
    $p_{mod} + n_{mod} - k$ & $p_{div} + 1$ & $n_{div} + 1$\\
    $k - n_{mod}$ & $p_{div} + 1$ & $n_{div}$\\
    $k - p_{mod}$ & $p_{div}$ & $n_{div} + 1$\\
    \bottomrule
    \end{tabular}
    }
    &
    \subfloat[][If $p_{mod} + n_{mod} \leq k$]{
    \begin{tabular}{ccc}
    \toprule
    count of folds & positives ($p_i$) & negatives ($n_i$) \\
    \midrule
    $k - p_{mod} - n_{mod}$ & $p_{div}$ & $n_{div}$\\
    $p_{mod}$ & $p_{div} + 1$ & $n_{div}$\\
    $n_{mod}$ & $p_{div}$ & $n_{div} + 1$\\
    \bottomrule
    \end{tabular}
   }
\end{tabular}
 \end{small}
 \end{center}
\end{table*}

    

\subsubsection{Testing in the lack of knowing the fold structure}
\label{sec:kfold}

When stratification is not used or its use is not indicated in a paper explicitly, any fold configuration can be assumed that satisfies the minimum requirements we formulated before. To address these scenarios, we note that if the number of folds is in the usual range (5-10) and the dataset is imbalanced or relatively small, it is feasible to enumerate all possible fold configurations and test each one of them. If all fold configurations are inconsistent with the reported scores, one can conclude that the experimental setup could not have yielded the reported scores. Although it might seem intractable to test all possible configurations in a real-life scenario, in Section \ref{sec:ehg}, we show real applications of this scenario. Enumerating all possible k-fold configurations given $p$, $n$ and $k$ is a non-trivial combinatorial problem. In the rest of this section, an algorithm is developed for this task.

The proposed algorithm is based on the observation that enumerating all fold configurations is closely related to the problem of integer partitioning in combinatorics and number theory \cite{intpart}. Specifically, we are interested in the various ways $p$ (or $n$) can be decomposed as $p = p_1 + \dots + p_k$. Given a particular partition, it be complemented with negatives to achieve the desired cardinality of folds ($n_i \sim (p + n)/k - p_i$), resulting in one fold configuration. There are algorithms proposed for the enumeration of all unique partitions of an integer to $m$ positive parts (for example, the algorithm on page 343 in \cite{intpart}). The only complexity that needs to be addressed is that the folds have varying cardinalities if the total number of elements ($p + n$) is not divisible by the number of folds ($k$).

Let $k_{div} = \lfloor (p + n) / k \rfloor$ and $k_{mod} = (p + n) \mod k$. There are two types of folds regarding cardinalities, that we denote by the superscripts $a$ and $b$: $k^a = k_{mod}$ folds, each with $c^{a} = k_{div} + 1$ elements, and $k^b = \defeq k - k_{mod}$ folds each with $c^b = k_{div}$ elements. Without the loss of generality, we choose the number of positives to drive the enumeration. Suppose there are $p^a$ positive samples in the folds of type $a$, implying $p^b = p - p^a$ positive samples in the folds of type $b$. The various configurations $p^a$ positives can be distributed in the folds of type $a$ can be determined by enumerating all integer partitions of $p^a$ into at most $k^a$ parts. Similarly, for folds of type $b$, one can determine all integer partitions of $p^b$ into at most $k^b$ parts. One combination of the partitions of positives in folds of type $a$ and $b$ can be complemented by negative samples to match the cardinality of the respective folds, resulting in a unique fold configuration. By iterating through all possible ways to split the total number of positives $p$ between the two types of folds, all fold configurations can be generated. A precise algorithm (with all technical details such as the removal of configurations not having a certain class present in at least two folds) is provided in Algorithm \ref{alg2}. We note that in practice, depending on the scores reported, further configurations can be removed, for instance, if sensitivity is reported, configurations with folds having zero positives could not drive the evaluation, since they lead to undefined sensitivity scores which contradicts the fact that the average sensitivity is reported.

\begin{algorithm}[t]
\caption{The algorithm generating all possible fold configurations (containing at least two folds with each class). $\mathbf{0}_{x}$ and $\mathbf{1}_{x}$ denote the $x$ dimensional 0-vector and 1-vector, respectively, $[\mathbf{x}, \mathbf{y}]$ stands for the concatenation of vectors $\mathbf{x}$ and $\mathbf{y}$.}\label{alg2}

\SetAlFnt{\small}
\SetKwFunction{FKFoldConfigurations}{KFoldConfigurations}
\SetKwProg{Fn}{Function}{:}{}
\SetKwProg{Gen}{Generator}{:}{}
\SetKwFunction{FMPartition}{MPartition}
\SetKwFunction{FAllPartitions}{Partition}
\SetKwFunction{FInvalidPCounts}{InvalidPCounts}
\SetKwFunction{FPRange}{PRange}
\SetKwFunction{FDistribution}{Distribution}
\SetKwFunction{FCombination}{Combinations}
\SetKwInOut{Yield}{Yield}
\SetKw{Args}{Args:}
\SetKw{Yields}{Yields:}
\SetKw{Description}{Description:}

\Gen{\FKFoldConfigurations{$p$, $n$, $k$}}{
\Description{Generates all fold configurations.}\\
\Args{the number of positives ($p$), negatives ($n$), and folds ($k\geq 2$)}\\
\Yields{$(\mathbf{p}, \mathbf{n}) \in \mathbb{N}^{k\times k}$, $(\mathbf{p}_i, \mathbf{n}_i)$ representing the number of positives and negatives in the $i$th fold, respectively.}

$k_{div}, k_{mod} \gets \lfloor (p + n) / k \rfloor, (p + n) \mod k$\;

$k^a, k^b \gets k_{mod}, k - k_{mod}$\;
$c^a, c^b \gets k_{div} + 1, k_{div}$\;

\For {$p^{a} = 0, \dots, \min\lbrace p, k^a\cdot c^a\rbrace$}{
    \For{$\mathbf{p}^a$ in \FAllPartitions{$p^a$, $k^a$, $c^a$}}{
        \For{$\mathbf{p}^b$ in \FAllPartitions{$p - p^a$, $k^b$, $c^b$}}{
            $\mathbf{p} \gets [\mathbf{p}^a, \mathbf{p}^b$]\;
            $\mathbf{n} \gets [\mathbf{1}_{k^a}\cdot c^a - \mathbf{p}^a, \mathbf{1}_{k^b}\cdot c^b - \mathbf{p}^b$]\;
            \If{$\sum\limits_{i=1}^{k} \mathbb{I}_{\mathbf{p}_i > 0} \geq 2 \wedge \sum\limits_{i=1}^{k}\mathbb{I}_{\mathbf{n}_i > 0} \geq 2$}{
            \Yield{$\mathbf{p}, \mathbf{n}$}
            }
        }
    }
}
}

\Gen{\FAllPartitions{$q$, $k$, $q_{max}$}}{
\Description{Generates all partitions of $q$ with at most $k$ parts, no part greater than $q_{max}$.}\\
\If{$q = 0$}{\Yield{$\mathbf{0}_k$}}
\For{$m = 1, \dots, \min(q, k)$ }{
\For{$\mathbf{q}$ in \FMPartition{q, m}}{
\If{$\sum\limits_{i=1}^{m}\mathbb{I}_{\mathbf{q}_i > q_{max}} = 0$}{
\Yield{$[\mathbf{0}_{k - m}, \mathbf{q}]$}
}
}
}
}

\Gen{\FMPartition{$q$, $m$}}{
\Description{Implements the algorithm on page 343 in \cite{intpart}, iteratively yielding one unique partitioning of $q$ to $m$ parts: $\mathbf{q}\in\mathbb{Z}_{+}^{m}$.}
}

\end{algorithm}

For instance, in the case of $p=30$, $n=300$, and $k=5$ (which is comparable in size to many small and imbalanced medical datasets), the total number of fold configurations is $673$. As an example, one particular output yielded by the generator in Algorithm \ref{alg2} is a pair of vectors $\mathbf{p}=[1, 2, 6, 9, 12], \mathbf{n} = [65, 64, 60, 57, 54]$ representing the fold configuration $[(p_1=1, n_1=65), (p_2=2, n_2=64), (p_3=6, n_3=60), (p_4=9, n_4=57), (p_5=12, n_5=54)]$.

\subsubsection{Testing in the lack of knowing the mode of aggregation}
\label{sec:lackagg}

If mode of aggregation (MoS or SoM) is unknown, one can still apply the proposed consistency tests to identify inconsistencies: the scores can be tested assuming both aggregations. If both tests lead to inconsistencies, one can safely conclude that the reported scores could not be the outcome of the experimental setup under either of the two reasonable modes of aggregation.

\section{Applications}
\label{sec:applications}

In this section, we illustrate the application of the proposed consistency tests in three real problems related to the use of machine learning in medicine, and also discuss potential further applications.

\subsection{Retinal vessel segmentation and further applications in retina image processing}
\label{sec:retina}

\begin{figure*}[t]
\begin{center}
\subfloat[(a)][Retina image\label{retimg}]{
\includegraphics[width=0.24\textwidth]{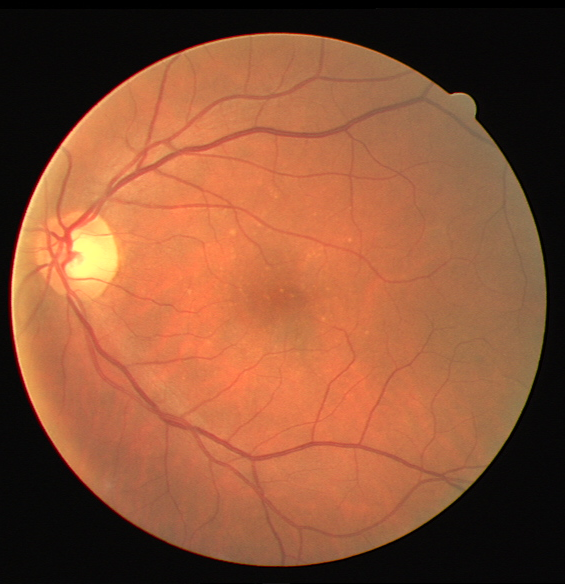}
}
\subfloat[(b)][Vessel mask\label{retves}]{
\includegraphics[width=0.24\textwidth]{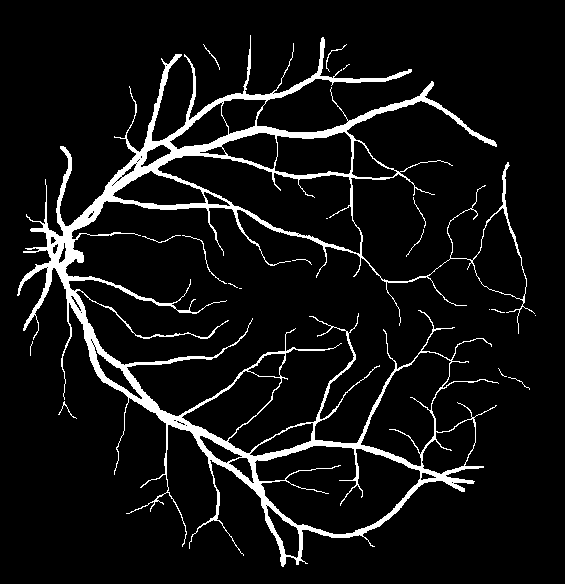}
}
\subfloat[(c)][Field of View mask\label{retfov}]{
\includegraphics[width=0.24\textwidth]{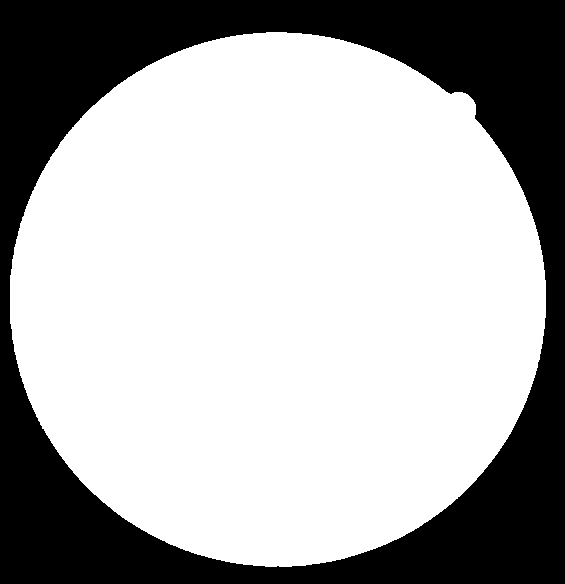}
}
\end{center}
\caption{One entry ("21") of the DRIVE retinal vessel segmentation dataset: a retina image (\ref{retimg}); the vessel mask (white pixels indicate the vasculature to be segmented) (\ref{retves}); the FoV mask, white pixels indicating the FoV region (\ref{retfov}).}
\label{fig:retina}
\end{figure*}

The techniques proposed in this paper are generalizations of the ones we used in our previous work \cite{vessel} in the field of retinal vessel segmentation. In this subsection, we provide a concise overview of that scenario, the results and potential applications in related fields. 
The field of retinal vessel segmentation has been a popular research area for nearly two decades. The most popular dataset for evaluation (DRIVE \cite{drive}) offers 20 training and 20 test images with manual annotations. Due to the well-defined test set, the reported performance scores (typically accuracy, sensitivity, and specificity) serve as the basis of ranking algorithms in most papers. 
Due to the specialities of the image acquisition techniques, the useful image content resides in a disk shaped area in the center of a rectangular image, referred as the \emph{Field of View} (FoV) (see Figure \ref{fig:retina} plotting one entry of the DRIVE dataset). Textual evidence suggested that some authors evaluate the performance of segmentation in the FoV region only, while others using all pixels of the images, but usually the region of evaluation is not specified explicitly. The problem is that the pixels outside the FoV region can easily be identified as non-vessel pixels, and account for about 30\% of all pixels: adding them to the true negatives boosts the accuracy and specificity scores compared to the case when the pixels covered by the FoV mask are used for evaluation.

Some authors report performance scores at the image level, and almost all authors provide average accuracy, sensitivity and specificity scores based on the 20 test images. In \cite{vessel}, we used specified versions of the tests proposed in Sections \ref{sec:ind}, \ref{sec:rom} and \ref{sec:mor} to assess the consistency of scores reported in 100 papers with two different assumptions: using only the pixels in the FoV region or using all pixels in the images for evaluation (note that the overall number of negatives $n$ is different). 
We found that about 30\% of the scores were inconsistent with any of the assumptions. The remaining scores were consistent with one of the assumptions and inconsistent with the other. A careful analysis of the rankings in the papers revealed that the rankings are based on incomparable scores in 100 papers. Beyond this insight, we also managed to derive a new baseline ranking by applying a correction to the scores derived from all pixels in the images to make them comparable with the ones calculated in the FoV region only.

Numerous further lesions and anatomical structures in retinal images, such as \emph{exudates} \cite{exu} and the \emph{optic disk} \cite{od}, are targeted by segmentation and detection algorithms. The possibility of evaluating their performance in the FoV region or using all pixels in the images is present in all related problems. Building on the successful application of the proposed techniques for vessel segmentation, it is reasonable to assume that these methods can be applied to validate and rectify the outcomes of other problems in retinal image processing.

\subsection{Preterm delivery prediction from electrohysterogram signals}
\label{sec:ehg}

In recent years, there has been growing interest in predicting preterm delivery from electrohysterogram (EHG) signals \cite{ehgreview}, particularly with the availability of the TPEHG dataset \cite{tpehg}, which comprises 38 positive and 262 negative records.
At some point, multiple authors reported almost perfect prediction scores in kFCV scenarios. However, in the study \cite{ehg}, it was revealed that these exceptionally high performance scores could not be replicated. The root cause of these overly optimistic results was traced to a methodological flaw: the improper use of minority oversampling.

The Synthetic Minority Oversampling TEchnique (SMOTE) \cite{smote} and its variations are commonly employed techniques to enhance the performance of binary classification on highly imbalanced data. These techniques involve artificially generating additional training samples for the minority class to address the asymmetric degeneracy in the learning process. When employing minority oversampling in a kFCV scenario, it is critical to apply oversampling to each training set separately, excluding any elements of the fold designated for testing. Applying oversampling to the entire dataset prior to kFCV adds highly correlated samples to the dataset, leading to a significant data leakage. The authors of \cite{ehg} reproduced all of the 11 papers and concluded that the most likely cause for the overly optimistic results is the application of minority oversampling prior to kFCV.

Since minority oversampling prior to kFCV increases the number of samples used for evaluation, one can expect inconsistencies between the reported scores and the correct experimental setup. Consequently, the cumbersome, time-consuming and error-prone work of reimplementing the algorithms could be replaced by employing the techniques developed in this paper. For example, one of the papers identified with the methodological flaw was \cite{ehgflaw2}, reporting $\hat{v}_{acc}^{MoS} = 0.9447$, $\hat{v}_{sens}^{MoS} = 0.9139$ and $\hat{v}_{spec}^{MoS} = 0.9733$ by 5-fold cross-validation. Although the authors mention that they used minority oversampling to increase the overall number of positives to $p'=244$, it is unclear if it was used only for training or for evaluation, as well.
The authors did not refer to using stratification, therefore, the evaluation of all fold configurations is needed for conclusive results. The number of possible fold configurations (with the correct $p=38$, $n=262$ counts and $k=5$) becomes 918 (by the exhaustive enumeration in Section \ref{sec:kfold}). The MoS test (Section {sec:mor}) reveals the inconsistency of scores with each configuration, leading to the conclusion that the reported scores are inconsistent with assumption of using 5-fold cross-validation on the original dataset. However, with the assumption of using $p'=244$ positive samples (that is, using the highly correlated generated samples for both training and evaluation), the overall number of possible fold configurations becomes $~2.6M$, but already the 962th configuration $[(1, 101), (4, 97), (40, 61), (99, 2), (100, 1)]$ serves as evidence that the reported scores could be yielded with $p'$ and $n$ counts, 5-fold cross validation and the actual $(tp_i, tn_i)$ counts $[(1, 96), (3, 92), (38, 59), (90, 2), (96, 1)]$. Consequently, the proposed method serves a numerical confirmation of the findings in \cite{ehg}: the reported scores could not be the outcome of a proper cross-validation, but they could be the outcome of the improper use of minority oversampling.

Beyond eliminating the need of reimplementation in this particular application, the example also highlights that the proposed techniques are suitable for detecting methodological flaws related to the increasingly prevalent use of minority oversampling in various fields \cite{smote}.

\subsection{Classification of skin lesions}
\label{sec:third}

In this section, we illustrate the application of the proposed techniques in a field where -- to our best knowledge -- no meta-analysis with the purpose of validating the reported results has been conducted before: the classification of skin lesion images. An analysis as detailed as the one we did in retinal vessel segmentation \cite{vessel} is clearly beyond the scope of the paper, however, we can test the reported scores in some highly influential papers to see if ambiguities are present. The analysis is based on the highly cited survey \cite{skinsurvey} providing a systematic overview of research up until 2021. From the survey, we have selected the 10 papers (listed in Table \ref{tab:skin}) with the most citations according to Google Scholar at the time of writing.

Unlike the problems in Sections \ref{sec:retina} and \ref{sec:ehg}, this field is centered around multiple datasets (see Table \ref{tab:skin}) with the task of classifying skin lesion images into two (malignant and non-malignant) or multiple, more specific classes. After a careful analysis of the selected papers, we found that \cite{skin1}, \cite{skin4} and \cite{skin6} are not suitable for testing (for the reason see the 'conclusion' column of Table \ref{tab:skin}). In the remaining papers, the authors share enough details to apply the consistency tests (the multiclass problem of ISIC2017 \cite{isic2017} was treated by the authors as two one-vs-all binary classification problems targeting the recognition of the classes \emph{melanoma} (M) and \emph{seborrheic keratosis} (SK)). In most papers, there are multiple sets of performance scores shared (illustrating the operation of certain steps of the algorithm), the most commonly reported ones being accuracy, sensitivity and specificity. Given that the datasets are either supplied with a test set of images, or the authors designate a test set and share its details, in each case the consistency test developed in Section \ref{sec:ind} was applied. The number of reported score sets ($N_{sc.}$) and the number of inconsistent sets ($N_{inc.}$) is shared in the corresponding columns of Table \ref{tab:skin}. In \cite{skin5} and \cite{skin8} no inconsistency was detected. In \cite{skin0}, \cite{skin2}, and \cite{skin3}, only a handful of score sets show inconsistencies, which might be caused by typos. However, the scores reported in papers \cite{skin7} and \cite{skin9} were found to be inconsistent with the assumption regarding the experiment, therefore, a more detailed analysis was carried out on these papers.

Regarding \cite{skin7}, we tested multiple assumptions, such as scores like $M_{ACC}$ represent the accuracy of the \emph{melanoma} (M) class against the \emph{nevus} class instead of both the \emph{nevus} and \emph{seborrheic keratosis} classes; and we also assumed that the accuracy and specificity figures are interchanged in the paper, since accuracy cannot be higher than both sensitivity and specificity the same time. All assumptions lead to inconsistencies with the claimed number and distribution of test images, therefore, we concluded that the reported scores are incomparable with the scores reported in other papers for the same dataset (such as \cite{skin3}). Regarding \cite{skin9}, the authors mention that they report the performance scores of binary classification in a weighted manner, from the perspective of both classes treated as positive and using the number of samples in the certain classes as weights. This uncommon weighting makes sensitivity the same as accuracy, and turns specificity into figure with no common interpretation. Although this reinterpretation resolves the inconsistencies, we still consider the scores inconsistent with the commonly accepted definitions of the terms.

The final conclusion of the analysis is that in the field of skin lesion classification, there are highly influential papers (such as \cite{skin7} and \cite{skin9}) with inconsistent scores reported and often recited for comparison and ranking in other research (e.g., \cite{skin7ref}).

\begin{figure*}[t]
\begin{center}
\subfloat[(a)][\emph{Nevus}\label{skinn}]{
\includegraphics[width=0.24\textwidth]{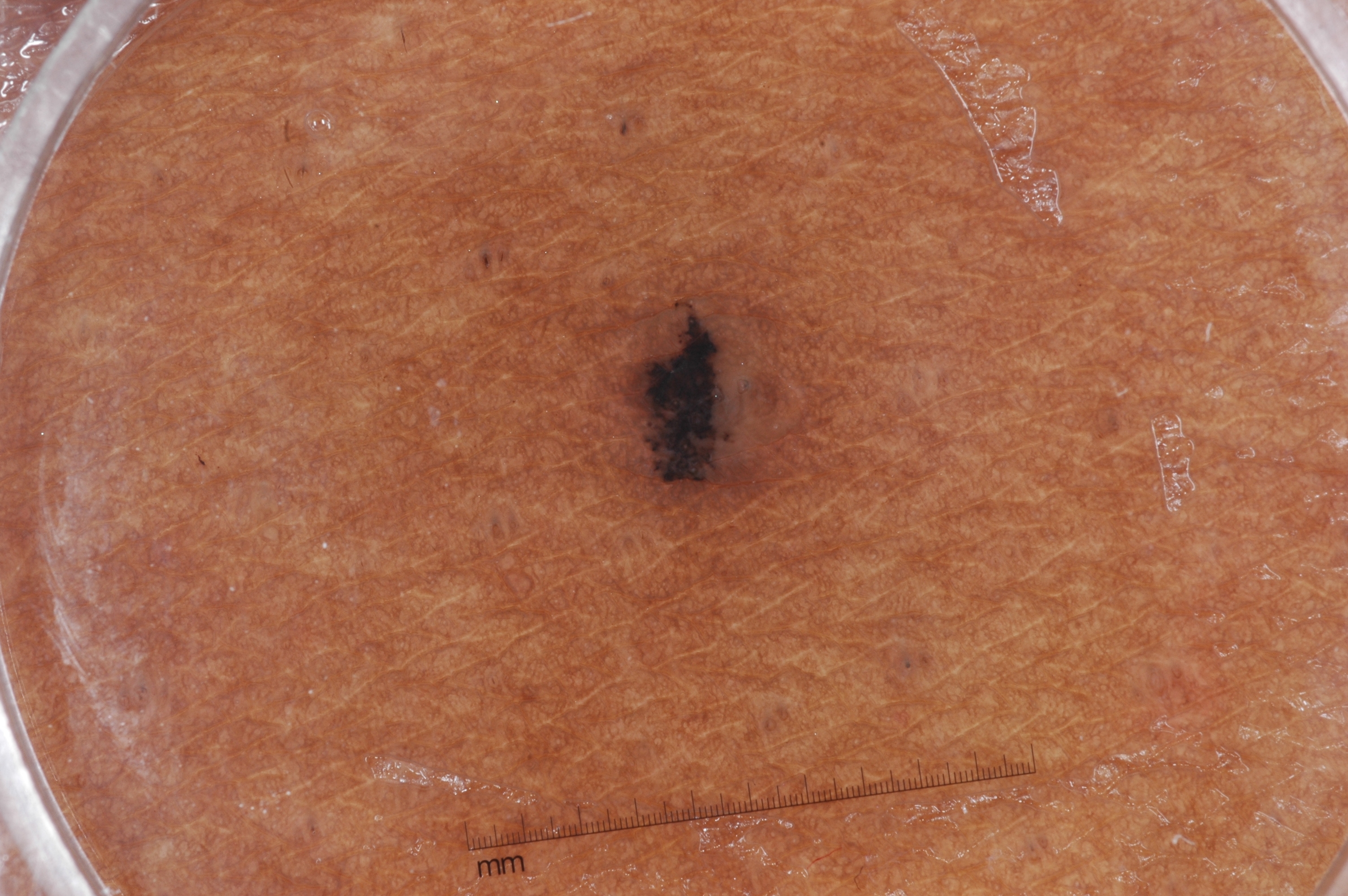}
}
\subfloat[(b)][\emph{Seborrheic keratosis}\label{skinsk}]{
\includegraphics[width=0.24\textwidth]{ISIC_0012134_sk.jpg}
}
\subfloat[(c)][\emph{Melanoma}\label{skinm}]{
\includegraphics[width=0.24\textwidth]{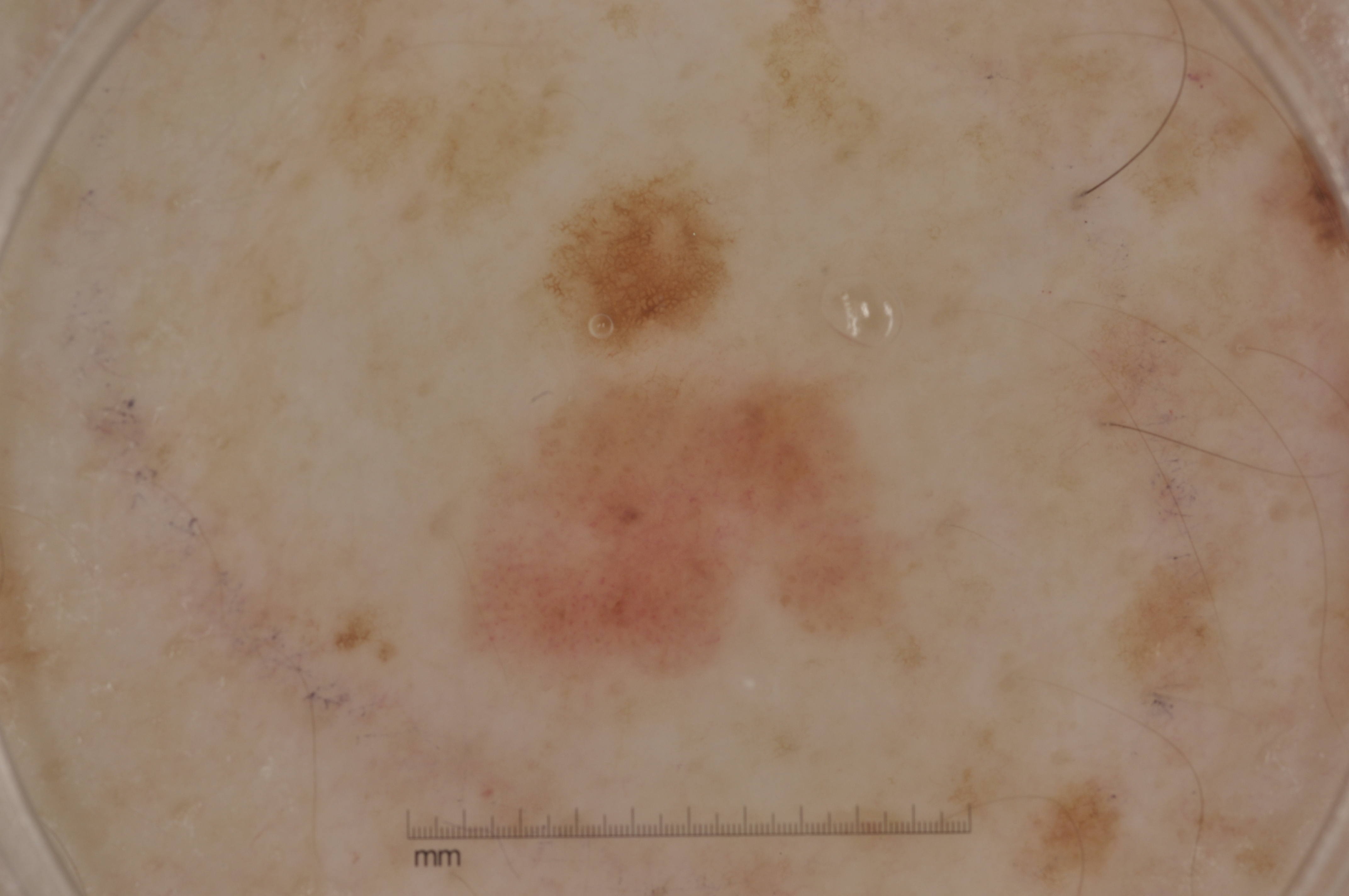}
}
\end{center}
\caption{Entries of the ISIC 2017 \cite{isic2017} dataset: \emph{Nevus} (\ref{skinn}); \emph{Seborrheic keratosis} (\ref{skinsk}); \emph{Melanoma} (\ref{skinm}).}
\label{figskin}
\end{figure*}

\begin{table*}
\caption{Summary of consistency checking in skin lesion classification}
\label{tab:skin}
\begin{center}
\begin{footnotesize}
\begin{tabular}{l@{\hspace{5pt}}r@{\hspace{5pt}}p{75pt}@{\hspace{5pt}}r@{\hspace{5pt}}p{75pt}@{\hspace{5pt}}l@{\hspace{5pt}}l@{\hspace{5pt}}p{200pt}@{\hspace{5pt}}}
\toprule
ref. & cit. & dataset & digits & scores & $N_{sc.}$ & $N_{inc.}$ & conclusion \\
\midrule
\cite{skin0} & 991 & ISIC2016 \cite{isic2016} & 3 & acc, sens, spec & 18 & 1 & Potentially typos present. \\
\cite{skin1} & 603 & ISIC2016 \cite{isic2016} & - & - & - & - & The paper is about image segmentation performance. \\
\cite{skin2} & 574 & ISIC2016 \cite{isic2016} & 3 & acc, sens, spec & 27 & 3 & Potentially typos present. \\
\cite{skin3} & 389 & ISIC2017 \cite{isic2017} m/sk & 3 & acc, sens, spec & 32 & 2 & Potentially typos present. \\
\cite{skin4} & 389 & ISIC2016 \cite{isic2016} (custom selection) & - & - & - & - & Not enough details shared. \\
\cite{skin5} & 322 & Argenziano's \cite{argenziano} & 3 & ppv, sens, spec & 16 & 0 & No inconsistency identified. \\
\cite{skin6} & 313 & custom & - & - & - & - & Not enough details shared. \\
\cite{skin7} & 312 & ISIC2017 \cite{isic2017} m/sk & 3 & acc, sens, spec & 20 & 20 & All accuracy, sensitivity and specificity scores reported for the two binary classification tasks M and SK are inconsistent. \\
\cite{skin8} & 259 & custom & 4 & acc, sens, spec & 18 & 0 & No inconsistency identified. \\
\cite{skin9} & 238 & ISIC2016 \cite{isic2016} & 4 & acc, sens, spec, f1 & 8 & 4 & Unorthodox weighting of the scores. \\
\bottomrule
\end{tabular}
\end{footnotesize}
\end{center}
\end{table*}

\section{Conclusions}
\label{sec:conclusions}

The meta-analysis of research is a crucial tool for addressing the reproducibility crisis in AI research and applications. Nevertheless, without numerical techniques, it demands enormous manual labor. To facilitate meta-analysis and enable the numerical identification of methodological inconsistencies, we have introduced various tests assessing the consistency of reported performance scores and experimental setups in binary classification. 

The proposed tests cover numerous evaluation scenarios. The test developed for performance scores derived from a single evaluation set supports 20 different scores (Section \ref{sec:ind}). We showed that the testing of scores aggregated by the \emph{Score of Means} strategy falls back to the case of scores derived from a single evaluation set (Section \ref{sec:rom}). For \emph{Mean of Scores} aggregations we developed a test supporting four of the most commonly reported scores (Section \ref{sec:mor}), and we also proposed the enumeration of all fold configurations when stratification is not used the specifics of the folds are unknown (Section \ref{sec:kfold}). Across Sections \ref{sec:ind} and \ref{sec:agg}, we highlighted multiple opportunities to improve the efficiency or extend the coverage of the tests.

In terms of applications, we briefly discussed the prior application of simplified forms of the proposed method in the field of retinal vessel segmentation (Section \ref{sec:retina}) and explored potential further applications related to retinal image processing. 
We also demonstrated that the proposed techniques are suitable for replacing the manual labor required for reimplementation in situations similar to the problem of preterm delivery prediction from EHG signals (Section \ref{sec:ehg}).
Additionally, this application illustrated that the proposed methods can be used to identify a common methodological pitfall when synthetic minority oversampling is employed.
Finally, in Section \ref{sec:third}, we revealed that irreproducible results are systematically recited in the literature of skin lesion classification, as well.

For the benefit of the community, a reference implementation of the proposed consistency tests was released as an open source Python package with an intuitive interface. The package is available in the standard Python repository (PyPI) under the name \verb|mlscorecheck| and on GitHub at the following URL: \url{https://github.com/FalseNegativeLab/mlscorecheck}.

\bibliographystyle{elsarticle-num}

\bibliography{references}

\end{document}